\documentclass[10pt,twocolumn,letterpaper]{article}

\usepackage[pagenumbers]{iccv} 

\definecolor{iccvblue}{rgb}{0.21,0.49,0.74}
\usepackage[pagebackref,breaklinks,colorlinks,allcolors=iccvblue]{hyperref}
\usepackage{multirow} 
\usepackage{amsmath} 
\usepackage{amssymb}  
\usepackage{marvosym}
\def\confName{ICCV}
\def\confYear{2025}

\title{Style4D-Bench: A Benchmark Suite for 4D Stylization}
\author{
Beiqi Chen$^{1,2^*}$ \quad
Shuai Shao$^{2^*}$ \quad
Haitang Feng$^{3,2}$ \quad
Jianhuang Lai${^4}$ \\
Jianlou Si$^{5\textsuperscript{\Letter}}$ \quad
Guangcong Wang$^{2\textsuperscript{\Letter}}$ \\
\vspace{0.4em} \\
$^1$Harbin Institute of Technology, \quad
$^2$Vision, Graphics, and X Group, Great Bay University, \\
$^3$Nanjing University, \quad
$^4$Sun Yat-Sen University, \quad
$^5$Alibaba Group \\ 
\href{https://becky-catherine.github.io/Style4D/}{Project page: https://becky-catherine.github.io/Style4D/}
}

\begin{document}
\maketitle
\begin{abstract}
We introduce \textbf{Style4D-Bench}, the first benchmark suite specifically designed for 4D stylization, with the goal of standardizing evaluation and facilitating progress in this emerging area. Style4D-Bench comprises: \textbf{1)} a comprehensive evaluation protocol measuring spatial fidelity, temporal coherence, and multi-view consistency through both perceptual and quantitative metrics, \textbf{2)} a strong baseline that make an initial attempt for 4D stylization, and \textbf{3)} a curated collection of high-resolution dynamic 4D scenes with diverse motions and complex backgrounds. To establish a strong baseline, we present \textbf{Style4D}, a novel framework built upon 4D Gaussian Splatting. It consists of three key components: a basic 4DGS scene representation to capture reliable geometry, a Style Gaussian Representation that leverages lightweight per-Gaussian MLPs for temporally and spatially aware appearance control, and a Holistic Geometry-Preserved Style Transfer module designed to enhance spatio-temporal consistency via contrastive coherence learning and structural content preservation. Extensive experiments on Style4D-Bench demonstrate that Style4D achieves state-of-the-art performance in 4D stylization, producing fine-grained stylistic details with stable temporal dynamics and consistent multi-view rendering. We expect Style4D-Bench to become a valuable resource for benchmarking and advancing research in stylized rendering of dynamic 3D scenes. 
\end{abstract}    
\section{Introduction}
\label{sec:intro}

Recent advances in 4D scene representations~\cite{wu20244d,luiten2024dynamic,duan20244d,yang2023real,pumarola2021d} have enabled high-fidelity modeling of dynamic environments, unlocking new possibilities for immersive content creation in virtual reality and film production. As applications become more demanding, there is a growing need not only for accurate reconstructions but also for controllable appearance and stylization of dynamic 4D content. Users may wish to stylize a 4D scene to reflect specific artistic intents, emotional tones, or narrative contexts—while maintaining both temporal coherence and multi-view consistency.

Despite recent progress in related areas like 2D and 3D stylization, the field of \textit{4D stylization} remains largely under-explored, with no standardized datasets, evaluation metrics, or task definitions. Existing methods adapted from 2D~\cite{gatys2016image,huang2017arbitrary,gu2018arbitrary,kolkin2019style,liao2017visual}, 3D~\cite{nguyen2022snerf,zhang2022arf,wang2023nerf,chiang2022stylizing,liu2024stylegaussian,saroha2024gaussian,zhang2024stylizedgs}, or video~\cite{lai2018learning,deng2021arbitrary} stylization often fail to meet the unique challenges of 4D settings, such as jointly preserving spatial fidelity, temporal stability, and multi-view consistency under complex motion and occlusion. Moreover, recent dynamic scene representations like 4D Gaussian Splatting (4DGS)~\cite{wu20244d} offer strong rendering capabilities, but have not yet been explored in the context of stylization.

To fill this gap, we introduce \textbf{Style4D-Bench}—the first benchmark suite dedicated to 4D stylization. Our benchmark provides: 1) A curated set of high-resolution, complex-background dynamic 4D scenes exhibiting diverse motions, deformations, and view-dependent effects; 2) A comprehensive evaluation protocol, including both perceptual and quantitative metrics, to assess spatial fidelity, temporal coherence, and cross-view consistency. Style4D-Bench is designed to facilitate fair, reproducible, and scalable evaluation of future 4D stylization approaches.

To establish a strong baseline within our benchmark, we also propose \textbf{Style4D}, a novel 4D stylization framework based on 4D Gaussian Splatting. Style4D consists of three key components: a basic 4DGS scene representation for geometry modeling, a \textit{Style Gaussian Representation} that incorporates lightweight per-Gaussian MLPs for time- and depth-aware stylization, and a \textit{Holistic Geometry-Preserved Style Transfer} module to enhance spatio-temporal consistency through contrastive learning and content-aware regularization.

Our contributions are summarized as follows: 1) We present \textbf{Style4D-Bench}, the first benchmark suite for 4D scene stylization, offering standardized datasets, tasks, and evaluation metrics to drive progress in this emerging area. 2) We define core evaluation challenges in 4D stylization, including spatial fidelity, temporal stability, and multi-view consistency, and design comprehensive protocols to quantify them. 3) We propose \textbf{Style4D} as a strong baseline method, which leverages 4D Gaussian Splatting with a style-aware representation and a holistic spatio-temporal transfer module. 4) Extensive experiments on Style4D-Bench demonstrate the effectiveness of our method and reveal insights into current limitations and future opportunities in 4D stylization.

\section{Related Works}
\label{sec:RW}

\label{gen_inst}

\textbf{3D Representations.} 
Neural Radiance Fields (NeRFs)~\cite{mildenhall2021nerf,barron2021mip} have transformed 3D scene representation by modeling scenes as continuous volumetric functions with MLPs, enabling high-quality novel view synthesis. Efficiency and quality have been further improved via compact representations such as decomposed tensors~\cite{chan2022efficient,chen2022tensorf,fridovich2023k}, hash tables~\cite{muller2022instant}, and voxel grids~\cite{fridovich2022plenoxels,sun2022direct}. Recently, 3D Gaussian Splatting (3DGS)~\cite{kerbl20233d} has emerged as a compelling alternative, representing scenes with explicit 3D Gaussians and enabling real-time rasterization-based rendering. Its explicit structure is particularly suitable for editing~\cite{chen2024gaussianeditor,haque2023instruct,liu2023stylerf,wang2022clip,zhuang2023dreameditor}, but 3DGS remains limited to static scenes.

\noindent\textbf{4D Representations.} To extend NeRF to dynamic settings, early works~\cite{park2021nerfies,pumarola2021d,park2021hypernerf} modeled temporal variations directly, while later methods~\cite{fang2022fast,liu2023robust,yi2023generalizable,cao2023hexplane,shao2023tensor4d} improved efficiency via voxel-based decompositions. Despite these advances, achieving real-time rendering and preserving fine-grained geometry under motion remains challenging. 4D Gaussian Splatting (4DGS)~\cite{wu20244d,duan20244d} addresses dynamic scenes by extending Gaussians into the temporal domain. Unlike Dynamic3DGS~\cite{luiten2024dynamic}, which stores per-frame parameters with linear memory growth, 4DGS uses a deformation network to model temporal changes compactly and efficiently. However, current 4D representations lack mechanisms for stylization and temporal consistency, motivating our benchmark and method for 4D scene stylization.

\noindent\textbf{2D Style Transfer.}  
Neural style transfer, pioneered by \cite{gatys2016image}, demonstrated that CNNs can effectively separate and recombine content and style from images. The key insight that second-order statistics of VGG features capture style information led to numerous improvements. Feed-forward methods like AdaIN \cite{huang2017arbitrary} significantly accelerated stylization by aligning feature statistics, while recent works have focused on improving semantic consistency and texture preservation \cite{gu2018arbitrary,kolkin2019style}. To maintain temporal coherence across frames, MCCNet \cite{deng2021arbitrary} proposes a Multi-Channel Correlation network. These foundational techniques form the basis for our 4D stylization component, which we enhance with temporal consistency mechanisms to handle dynamic scenes.

\noindent\textbf{3D and Video Stylization.} Extending style transfer to 3D has attracted significant attention. NeRF-based methods achieve multi-view consistent stylization through optimization \cite{nguyen2022snerf,zhang2022arf,wang2023nerf} or feed-forward networks \cite{liu2021adaattn,chiang2022stylizing}. With the advent of 3D Gaussian Splatting, StyleGaussian \cite{liu2024stylegaussian} demonstrates instant style transfer by embedding VGG features into Gaussians and aligning their statistics with style images. GSS \cite{saroha2024gaussian} and StylizedGS \cite{zhang2024stylizedgs} also leverage 3DGS but require per-style optimization.  While these methods excel at static scenes, they fail to handle temporal dynamics inherent in 4D content, such as motion, deformation, and occlusion, leading to view inconsistencies and temporal flickering. Existing video stylization methods~\cite{chen2024upst,huang2022stylizednerf,wang2023nerf} primarily focus on temporal coherence in 2D space without incorporating 3D scene understanding, whereas 3D stylization methods typically disregard temporal dynamics altogether.

\noindent\textbf{4D Stylization and Benchmark.} Despite notable progress in 2D and 3D stylization, \textit{4D scene stylization} remains largely unexplored. The core challenge lies in simultaneously ensuring multi-view consistency across different viewpoints and temporal stability across frames. Recent concurrent efforts such as 4DStyleGaussian~\cite{liang20244dstylegaussian} make early steps toward 4D stylization, yet they still face limitations in content preservation, geometric fidelity, and temporally adaptive style control. Our strong baseline addresses these challenges by introducing per-Gaussian MLPs to enable fine-grained, temporally-aware appearance modulation, and by integrating an enhanced 2D stylization module designed to preserve spatial-temporal consistency throughout the 4D stylization process.

Designing a benchmark for 4D stylization poses fundamental challenges. The absence of ground truth and the subjective nature of style make quantitative evaluation inherently difficult. Moreover, measuring temporal coherence and multi-view consistency is non-trivial due to complex spatial-temporal dynamics and the lack of reference sequences. Existing evaluation protocols from 2D, 3D, or video stylization are insufficient for capturing these aspects, highlighting the need for dedicated datasets and tailored metrics.
\section{Style4D-Bench}
\label{bench}


Existing 4D stylization tasks lack a unified quantitative evaluation standard. Some studies \cite{liang20244dstylegaussian} measure consistency by calculating frame-wise PSNR, MSE, and other metrics, using style loss to assess stylization degree. In contrast, others \cite{liu2021adaattn} overly rely on user studies. While these metrics simplify the evaluation process, they introduce several issues. First, simple style losses like MSE overly focus on pixel-level differences between stylized and reference images, failing to accurately reflect overall similarity in semantic and textural aspects. Second, 4D stylization is a complex and multidimensional concept where individual preferences may prioritize different aspects. For instance, some emphasize stylization degree, considering blurriness of characters or objects as part of style, while others prioritize preserving the structure and details of the original scene. To address these challenges, we propose a decomposition method that breaks down the evaluation of 4D stylization into multiple dimensions for a more comprehensive and nuanced assessment.

We divide the evaluation of 4D stylization into two aspects: 4D stylization quality and 4D stylization consistency. For 4D stylization quality, we focus on assessing the frame-by-frame quality of stylized videos rendered at arbitrary times and angles, without considering similarity to a reference style. Regarding 4D stylization consistency, we distinguish between Temporal Quality and Stylization Quality. Our evaluation method encompasses a total of six specific dimensions, comprising 12 detailed metrics in total.

\subsection{4D Stylization Quality}

\noindent\textbf{Frame-Wise Quality - Imaging Quality.}
Imaging quality refers to the distortions (e.g., blurring, high noise, overexposure) present in each frame after 4D stylization. We assess this quality using three metrics: UIQM, Clipiqa+\cite{wang2022exploring}, and Musiq\cite{ke2021musiq}.

\begin{itemize}
\item UIQM: Integrates three dimensions of image quality—color, sharpness, and contrast—and computes the overall image quality through a weighted averaging approach, aligning with human visual perception.
\item Clipiqa+: A CLIP-based image quality assessor fine-tuned using CoOp \cite{zhou2022extract}, designed to evaluate overall image quality.
\item Musiq: A multi-scale image quality assessor trained on the KonIQ dataset \cite{hosu2020koniq}, capable of capturing and evaluating image quality at various granularities.
\end{itemize}

\noindent\textbf{Frame-Wise Quality - Aesthetic Quality.}
Aesthetic quality refers to the artistic and aesthetic value of each video frame, reflecting aspects such as layout, color richness and harmony, photorealism, naturalness, and artistic quality. We assess aesthetic quality using the Qalign\cite{wu2023qalign} and Musiq-paq2piq\cite{ke2021musiq} metrics.
\begin{itemize}
\item Qalign: A pre-trained large multimodal model (LLM) capable of simultaneously assessing both image quality and aesthetic performance.
\item Musiq-paq2piq: A multi-scale image aesthetic assessor trained on the PaQ-2-PiQ dataset\cite{ying2020patches}, capable of capturing image quality and performing aesthetic evaluation at various granularities.
\end{itemize}

\subsection{4D Stylization Consistency}
\noindent\textbf{Temporal Quality - Spatiotemporal Consistency.} Temporal consistency refers to maintaining continuity between frames and consistency across multiple viewpoints as time and perspectives change. We evaluate multiple helical trajectory videos rendered for each scene using Dists\cite{ding2020iqa} and Warp loss metrics to assess cross-frame consistency.

\begin{itemize}
\item Dists: Comprehensively measures both the structural and textural similarity between two images to evaluate their overall quality and perceptual consistency.
\item Warp loss: Using RAFT\cite{teed2020raft} to compute optical flow, mapping the next frame to the current frame, and calculating the L1 error between the current frame and the mapped frame effectively evaluates spatial and motion consistency of images in a temporal sequence
\end{itemize}

\noindent\textbf{Temporal Quality - Subject Consistency.} For stylized videos rendered from a fixed viewpoint, we assess whether the appearance of a subject (e.g., a person, curtains, etc.) remains consistent throughout the entire video. To achieve this, we compute the inter-frame DINO\cite{caron2021emerging} feature similarity.

\noindent\textbf{Stylization Quality - Style Consistency.}Style consistency refers to the degree of stylization in 4D stylization, evaluated using CKDN\cite{zheng2021learning} and LPIPS\cite{zhang2018unreasonable} metrics to assess the similarity between video frames and style images, representing the level of stylization.
\begin{itemize}
\item CKDN: Utilizes learned representations from degraded images to assess style similarity, enabling comprehensive evaluation of both image quality and style similarity.
\item LPIPS: A metric based on features extracted from VGG, measuring perceptual similarity between images, effectively assessing similarity to style images.
\end{itemize}

\noindent\textbf{Stylization Quality - Content Consistency.} Content Consistency refers to the semantic and content similarity between stylized 4D scenes and their original counterparts. We employ SSIM and LPIPS metrics to compute the similarity of each frame to the original scene.

\subsection{User Study Design}
To thoroughly evaluate our method, we conducted a user study with a total of 34 participants. The study consists of two parts. The first part focuses on 4D stylization, where we presented five pairs of videos—each approximately 10 seconds long and containing 300 frames—from test viewpoints and arbitrary transformed viewpoints, along with four selected frames extracted for detailed qualitative assessment. The second part aims to validate the effectiveness of our proposed HGST method in video stylization, featuring two pairs of 10-second videos (300 frames each) as well as six individually selected frames for fine-grained evaluation.

\textbf{Metric.} For the extracted single frames, we evaluated two metrics: stylization quality and image quality. Stylization quality measures the extent to which edges and textures are transformed to reflect the target style without compromising the original image structure. Image quality assesses the clarity of object boundaries and facial details. 

\begin{figure*}[h]
  \centering
  \includegraphics[width=1\textwidth]{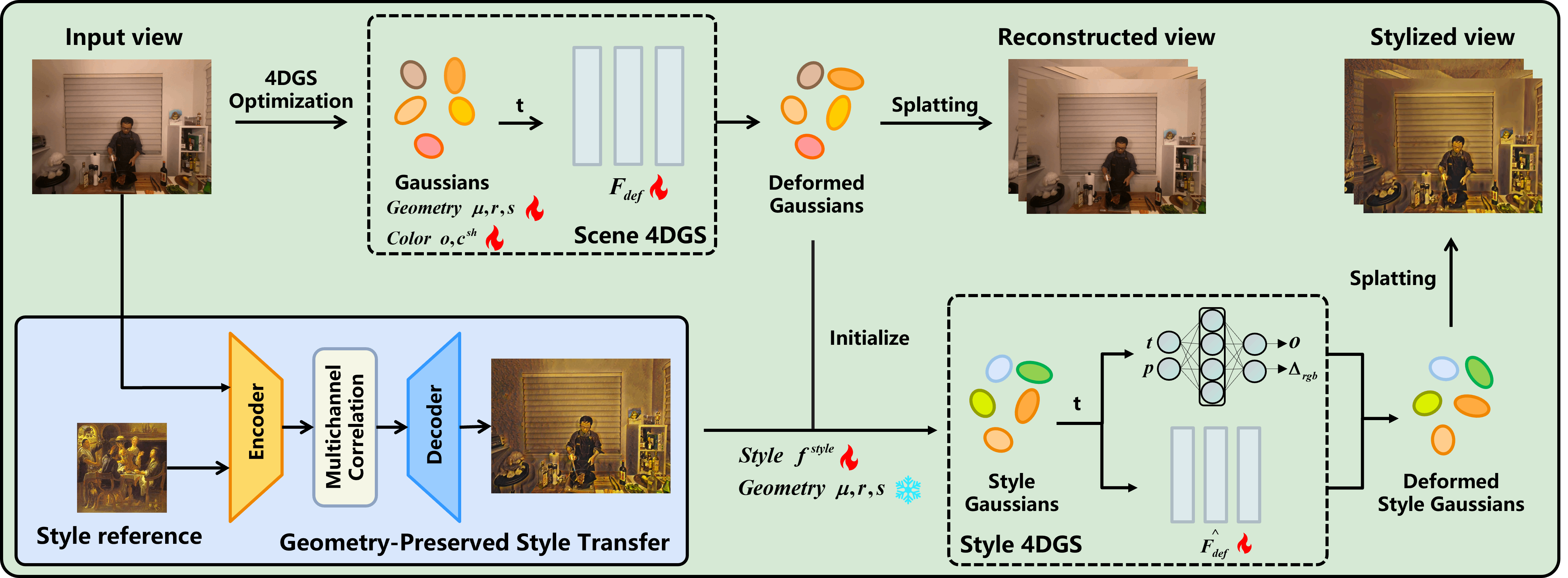}
  \caption{Overview of Style4D. Style4D consists of three key components, a basic 4DGS representation, a Style Gaussian Representation, and a Holistic Geometry-preserved Style Transfer. We first train a basic 4DGS representation with the content image to obtain 4D scene geometry. Then we propose a new Style Gaussian Representation for 4D stylization. We also introduce a Holistic Geometry-preserved Style Transfer module to improve consistency and quality of stylization.}
  \label{fig:styleframework}
\end{figure*}

For the continuous long videos, we adopted three metrics: stylization quality, spatiotemporal consistency, and video quality. Stylization quality evaluates how well the video’s style matches the reference while preserving the original structure. Spatiotemporal consistency measures the coherence of the video across temporal progression and viewpoint changes. Video quality assesses the clarity of fine details and textures, as well as overall visual preference.

\section{Style4D: A Strong Baseline}
\label{headings}

\noindent\textbf{Overview of Style4D.} In 4D stylization, the direct use of style transformation leads to low multi-view consistency and significant blurry artifacts. To address these issues, we propose Style4D, a new dynamic scene stylization framework, based on the decoupling of geometry learning and style learning. The framework of Style4D is illustrated in Figure \ref{fig:styleframework}. Style4D consists of three key components, a basic 4DGS representation, a Style Gaussian Representation, and a Holistic Geometry-preserved Style Transfer. We first train a basic 4DGS representation given multiple views $ I_{\mathrm{content}}$ to obtain the static Gaussian sequence $ \mathcal{G}_i = \{\boldsymbol{\mu}_i, \boldsymbol{r}_i, \boldsymbol{s}_i, o_i, c_{i}^{\mathrm{sh}}\} $ and the corresponding Gaussian Deformation Field Network $ F_{\mathrm{def}}$, capturing the the geometry of a dynamic scene. To stylize a 4D scene represented by 4DGS, we propose a Style Gaussian Representation method. It is a novel type of Gaussian ellipsoid, with attributes defined as $ \mathcal{G}_i = \left\{ \boldsymbol{\mu}_i, \boldsymbol{r}_i, \boldsymbol{s}_i, \mathbf{f}_i^{\mathrm{style}} \right\} $. We design a tiny MLP as part of the style attribute $ f_i^{\mathrm{style}}$. The design provides pixel-level stylization mapping, and thus achieves finer color expression while balancing local and global consistency, which significantly improves multi-view consistency. Finally, we introduce a geometry-preserving style transfer approach that integrates an attention-guided 2D stylization module with contrastive coherence learning, enabling the generation of high-quality and temporally consistent training frames.

\noindent\textbf{Style Gaussian Representation.} Inspired by SuperGaussians~\cite{xu2024supergaussiansenhancinggaussiansplatting}, which enhances 2DGS using bilinear interpolation and spatially varying features, we introduce a \textit{Style Gaussian Representation} for 4D scenes. 
Extending kernel functions to four-dimensional interpolation does not effectively capture the four-dimensional variations in scenes. Therefore, we introduce Gaussian MLP features. It is worth noting that directly mapping intersection coordinates to color and opacity can lead to overfitting. Thus, we map color and opacity variations based on intersection depth.
Specifically, each Gaussian is assigned a tiny MLP and a style code $f^{\mathrm{style}}$ to modulate color and opacity over space and time. Given the camera pose $M = [R, T]$, we compute the ray-ellipsoid intersection point $p_t$ from pixel $p$ and time $t$, following~\cite{yu2024gaussian}. The color is then rendered as:
\begin{equation}
\label{eq:style_splating}
c(p) = \sum_{i \in \mathcal{N}(p)} (c_i + F_c(p_t, t)) \cdot F_\alpha^i \prod_{j=1}^{i-1}(1 - \alpha_j(p)),
\end{equation}
where $c_i$ is the view-dependent base color, $F_c(p_t, t)$ is the style-driven color increment from the MLP, and $\alpha_j(p)$ denotes opacity. All $F_c$ and $\alpha_j$ terms are predicted per-Gaussian via MLPs, with $p_t$ as the ray-Gaussian intersection depth. This formulation preserves 3D geometry, allows precise temporal control, and enhances multi-view consistency.

\noindent\textbf{Holistic Geometry-preserved Style Transfer.} Long video stylization remains challenging, especially for high-resolution sequences. Diffusion-based methods~\cite{feng2024ccedit,kara2024rave,yang2024cogvideox} often suffer from structural distortions and temporal inconsistency. Optical flow-based constraints~\cite{liu2021structure,chen2017coherent} improve coherence but are computationally expensive and scale poorly. Self-supervised approaches~\cite{10008203,wu2022ccpl} reduce flickering but may introduce artifacts such as hollow textures and sharp pixel boundaries due to lack of semantic guidance.

\begin{table*}[h]
\centering
\resizebox{\textwidth}{!}{
\begin{tabular}{cccccccccccccc}
\hline
\multirow{2}{*}{Method} & \multirow{2}{*}{Dataset}          & \multicolumn{3}{c}{Imaging Quality}                  & \multicolumn{2}{c}{Aesthetic Quality} & \multicolumn{2}{c}{Spatiotemporal Consistency} & Subject Consistency & \multicolumn{2}{c}{Style Consistency} & \multicolumn{2}{c}{Content Consistency} \\ \cline{3-14} 
                        &                                   & UIQM↑           & Clipiqa+↑       & Musiq↑           & Qalign↑           & Musiq-paq2piq↑    & Dists↓                 & Warp Loss↓            & DINO Score↑         & CKDN↑             & LPIPS↓            & SSIM↑              & LPIPS↓             \\ \hline
4DGS(AdaIN)             & \multirow{4}{*}{cook\_spinach}    & 1.2995          & 0.4209          & 49.4095          & 2.8665            & 60.2165           & 0.0114                 & 0.0091                & 0.9309              & 0.2084            & 0.6913            & 0.4763             & 0.4898             \\
4DGS(AdaAttN)           &                                   & 1.7240          & 0.3770          & 36.1325          & 2.1298            & 50.7028           & 0.0215                 & 0.0117                & 0.9078              & 0.2173            & 0.6904            & 0.6444             & 0.2841             \\
4DStyleGaussian         &                                   & 1.0834          & 0.4267          & 44.2200          & 2.7019            & 55.2280           & 0.0121                 & \textbf{0.0053}       & \textbf{0.9403}     & 0.1978            & 0.7106            & 0.7646             & 0.2159             \\
Style4D(Ours)           &                                   & \textbf{1.9290} & \textbf{0.4437} & \textbf{53.4681} & \textbf{3.2072}   & \textbf{65.8520}  & \textbf{0.0112}        & 0.0058                & 0.9395              & \textbf{0.2290}   & \textbf{0.6866}   & \textbf{0.7771}    & \textbf{0.1834}    \\ \hline
4DGS(AdaIN)             & \multirow{4}{*}{flame\_salmon\_1} & 1.6918          & 0.3336          & 41.3119          & 2.8497            & 51.6363           & 0.0171                 & 0.0141                & 0.9267              & 0.2193            & 0.7605            & 0.5201             & 0.4030             \\
4DGS(AdaAttN)           &                                   & 1.2928          & 0.3120          & 39.9117          & 2.2357            & 54.5593           & 0.0271                 & 0.0169                & 0.9072              & 0.1966            & 0.7861            & 0.6054             & 0.2944             \\
4DStyleGaussian         &                                   & 1.5488          & 0.3218          & 51.3875          & 3.2182            & 61.4087           & 0.0140                 & 0.0074                & 0.9402              & 0.1897            & 0.7701            & 0.5081             & 0.6051             \\
Style4D(Ours)           &                                   & \textbf{1.7529} & \textbf{0.3962} & \textbf{55.2302} & \textbf{3.6030}   & \textbf{63.5178}  & \textbf{0.0138}        & \textbf{0.0067}       & \textbf{0.9415}     & \textbf{0.2354}   & \textbf{0.7602}   & \textbf{0.6963}    & \textbf{0.2704}    \\ \hline
4DGS(AdaIN)             & \multirow{4}{*}{sear\_steak}      & 1.3544          & 0.3430          & 51.6330          & 2.6115            & 62.2037           & 0.0129                 & 0.0078                & 0.9463              & 0.2352            & 0.7114            & 0.6000             & 0.2987             \\
4DGS(AdaAttN)           &                                   & 1.1910          & 0.3996          & 43.6474          & 2.0062            & 63.0785           & 0.0234                 & 0.0126                & 0.9021              & 0.3204            & 0.7050            & 0.5153             & 0.4768             \\
4DStyleGaussian         &                                   & 1.3843          & 0.3613          & 42.1204          & 2.5841            & 56.5093           & 0.0131                 & \textbf{0.0050}       & 0.9530              & 0.2722            & 0.7161            & 0.6557             & 0.4819             \\
Style4D(Ours)           &                                   & \textbf{1.6818} & \textbf{0.4176} & \textbf{53.3443} & \textbf{2.8820}   & \textbf{68.0488}  & \textbf{0.0108}        & 0.0066                & \textbf{0.9564}     & \textbf{0.3239}   & \textbf{0.7014}   & \textbf{0.7503}    & \textbf{0.2146}    \\ \hline
\end{tabular}}
\caption{Quantitative comparisons of our proposed Style4D against state-of-the-art methods on Style4D-Bench.}
\label{tab:Quantitative_table}
\end{table*}

To address these issues, we propose a \textit{Holistic Geometry-preserved Style Transfer} (HGST) module based on an encoder-transformer-decoder architecture~\cite{10008203}. We fuse style and content features using Multichannel Correlation to ensure global consistency, and decode stylized outputs with better structural integrity. However, the encoder-decoder pipeline still leads to temporal instability, particularly on large unseen frames. Inspired by~\cite{wu2022ccpl}, we introduce a dual constraint: an attention-guided local contrastive loss $\mathcal{L}_{lcl}$ and a global feature consistency loss $\mathcal{L}_{content}$, which jointly enhance local coherence and global structure, effectively mitigating flickering and spatial artifacts. 
To enhance local and global coherence, we extract multi-scale features from $I_{cs}$, $I_{content}$, and $I_{style}$ via an encoder, denoted as $f_i^{cs}$, $f_i^c$, and $f_i^s$ for layers $i=1$ to $5$. For $i=3$ to $5$, we apply CBAM~\cite{CBAM} to enhance salient features and randomly sample $N$ locations $G_x$ and their 8-nearest neighbors $G_{x,y}$ with small perturbations. Local differences are defined as $d_{x,y}^g = G_x^{f_i^{cs}} - G_{x,y}^{f_i^{cs}}$, $d_{x,y}^c = G_x^{f_i^{c}} - G_{x,y}^{f_i^{c}}$. We employ a contrastive loss to maximize similarity between aligned local differences:
\begin{equation}
\mathcal{L}_{lcl} = \sum_{m=1}^{8N} - \log \frac{\exp(d_m^g \cdot d_m^c / \tau)}{\sum_{n=1}^{8N} \exp(d_m^g \cdot d_n^c / \tau)}, \quad \tau=0.07.
\end{equation}

To mitigate artifacts and preserve global structure, we introduce a global content loss:
\begin{equation}
\mathcal{L}_{content} = \frac{1}{N} \sum_{i=1}^N \| f_{csi} - f_{ci} \|_2^2.
\end{equation}

The final consistency loss combines both terms:
\begin{equation}\label{eq:consistency}
\mathcal{L}_{consistency} = \mathcal{L}_{lcl} + \mathcal{L}_{content}.
\end{equation}

\noindent\textbf{Training Objective.} We first train a 4D Gaussian Splatting model using multi-view content images $I_{content}$ to reconstruct the scene geometry, yielding a static Gaussian sequence $\mathcal{G}_i$ and a deformation field network $F_{def}$. We then train a holistic geometry-preserved style transfer module with the following overall objective:
\begin{equation}
\begin{aligned}
\mathcal{L}_{total} =\ & \lambda_{consistency} \mathcal{L}_{consistency} + \lambda_{style} \mathcal{L}_{style} \\
& + \lambda_{id} \mathcal{L}_{id} + \lambda_{illum} \mathcal{L}_{illum} + \lambda_{ins} \mathcal{L}_{ins},
\end{aligned}
\end{equation}
where $\mathcal{L}_{consistency}$ ensures spatio-temporal coherence (see Eq.~\ref{eq:consistency}), and the remaining terms follow conventional stylization objectives~\cite{10008203}: $\mathcal{L}_{style}$ for perceptual style alignment, $\mathcal{L}_{id}$ for content preservation, $\mathcal{L}_{illum}$ for illumination stability, and $\mathcal{L}_{ins}$ for intra-channel coherence. See supplementary material for definitions. After obtaining high-quality stylized frames, we train a Style Gaussian Representation supervised by these frames. The optimization objective is:
\begin{equation}
\mathcal{L} = \left\| \hat{I} - S_t(I) \right\|_1 + \mathcal{L}_{tv},
\end{equation}
where $\hat{I}$ is the rendered image, $S_t(I)$ is the corresponding stylized frame, and $\mathcal{L}_{tv}$ denotes total variation regularization for spatial smoothness.


\begin{figure*}[h]
  \centering
  \includegraphics[width=0.95\textwidth]{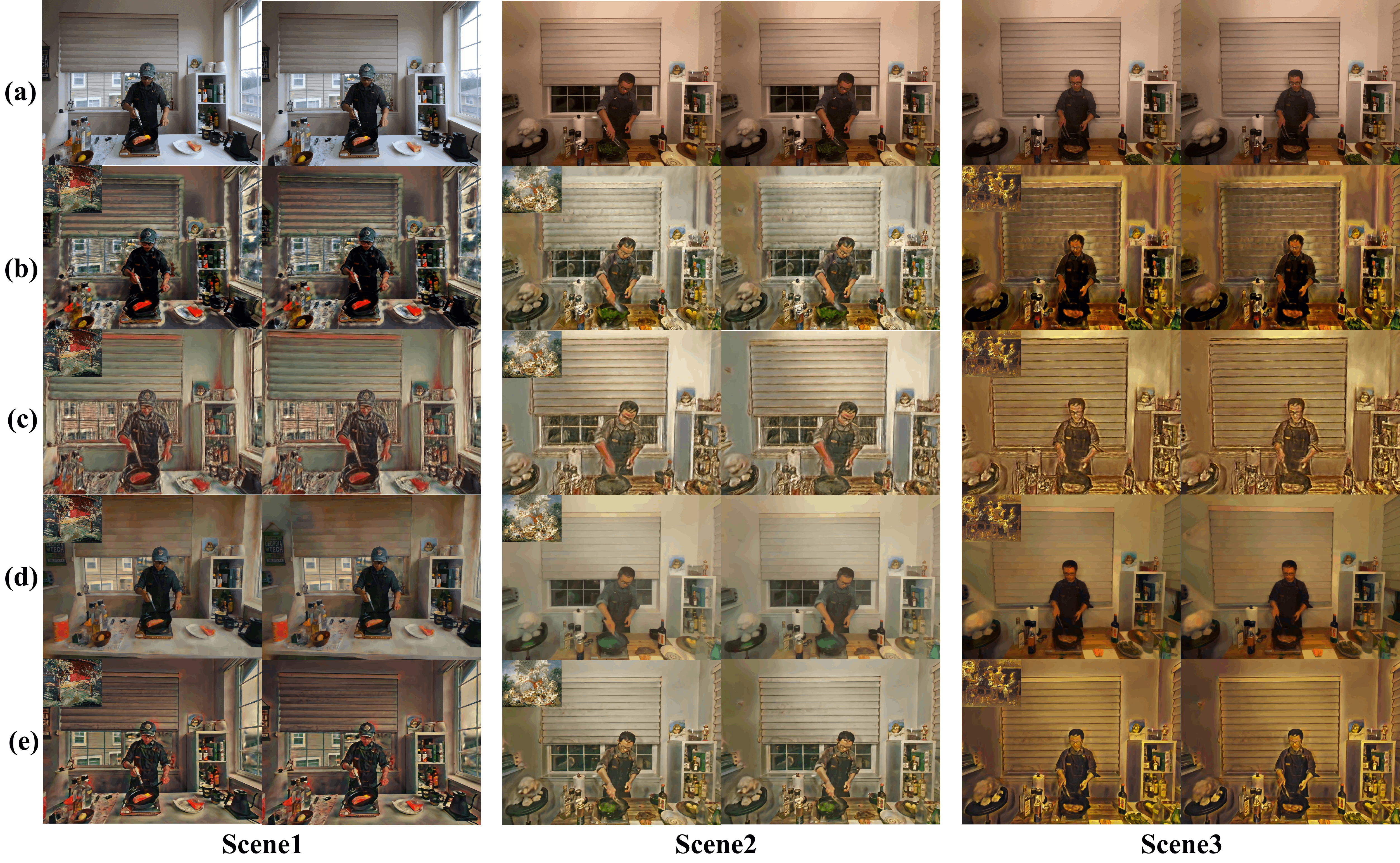}
  \caption{4D Stylization Comparison: (a) Original scene image, (b) 4DGS with AdaIN, (c) 4DGS with AdaAttN, (d) 4DStyleGaussian, (e) Style4D (Ours). Please refer to supplementary material for rendered videos.}
  \label{fig:Qualitative_style4d}
\end{figure*}

\section{Experiments}
\label{sec:experiments}

\textbf{Datasets.} We evaluate our model on the real-world Neu3D dataset \cite{li2022neural} to benchmark its performance in realistic scenarios. Neu3D comprises six dynamic scenes, each observed by 15–20 static cameras distributed in space. The dataset features long video sequences (300 frames), with complex scene dynamics and nontrivial viewpoint variations. The videos are recorded at a resolution of 1352 × 1014 pixels, with 300 frames per sequence.

\noindent\textbf{Experiment settings.}
We build our implementation of Style4D based on the publicly available 4DGS codebase \cite{wu20244d}. During training, we adopt the Adam optimizer with hyperparameters following those used in 4DGS. The batch size is set to 2, and each lightweight per-Gaussian MLP consists of two layers. For each scene, we train the model for up to 14,000 iterations. Both training and inference are performed on a single NVIDIA A40 GPU with 48GB of memory. In practice, training a single scene takes approximately 2 hours, with peak GPU memory usage around 10GB.

\subsection{Evaluation on Style4D-Bench}

\textbf{Qualitative results.} We compare our method with existing 4D stylization methods including 4DStyleGaussian, as well as baseline 4DGS models trained with AdaIN and AdaAttN stylized images, to assess stylization quality. As shown in Figure \ref{fig:Qualitative_style4d}, our method demonstrates stronger temporal consistency compared to 4DGS with AdaIN and AdaAttN, exhibiting fewer artifacts and blurriness on background objects. Moreover, compared to 4DStyleGaussian, our approach significantly enhances stylization effects while maintaining consistency and effectively preserving structural details of the original scenes without excessive smoothing. Due to space limitations, we only present results from fixed test viewpoints here. Stylization results from spiral viewpoints are provided in the appendix, where our method demonstrates stronger consistency and enhanced stylization quality.

\begin{figure*}[h]
  \centering
  \includegraphics[width=0.95\textwidth]{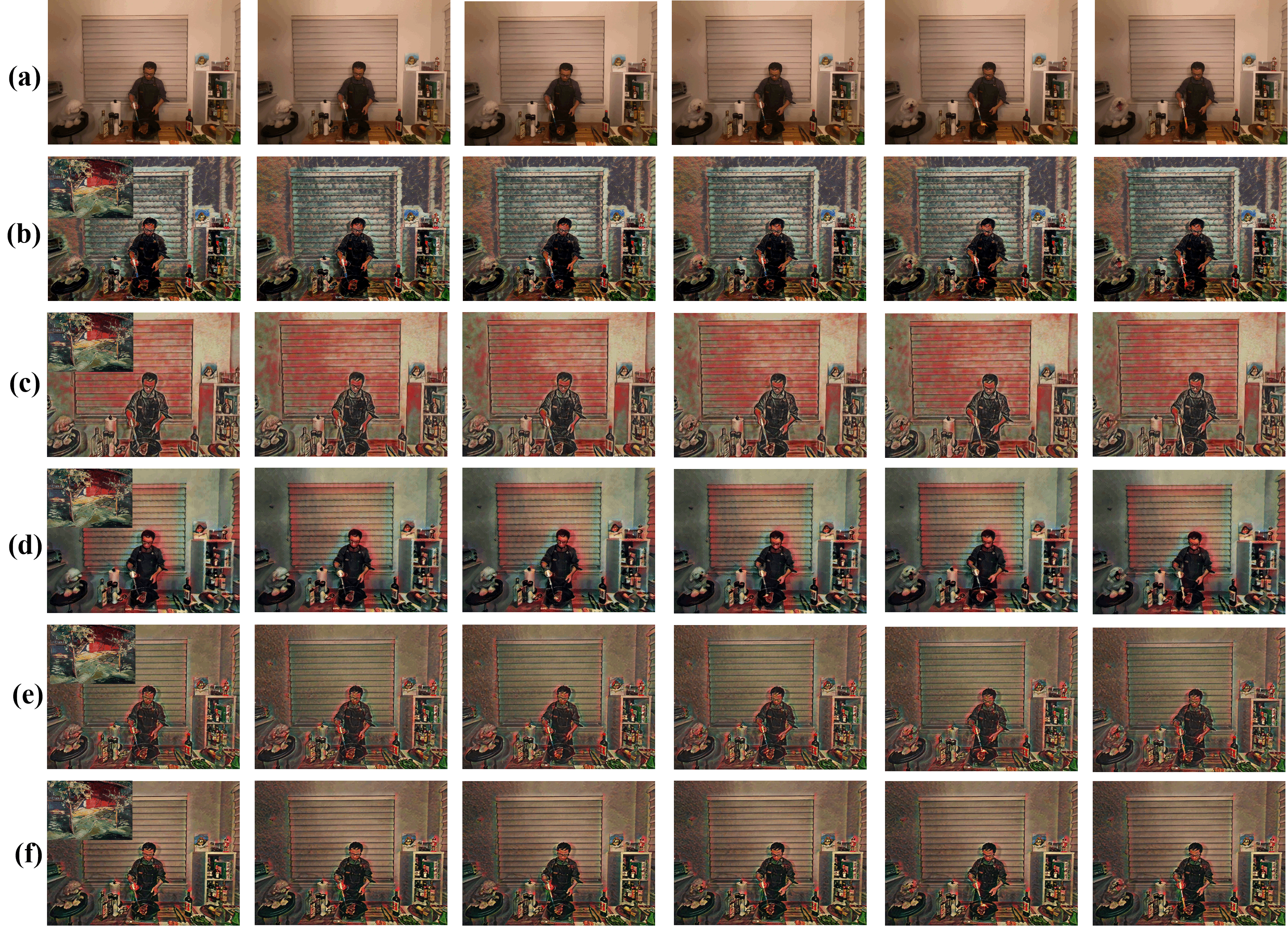}
\caption{Visualization of HGST stylization method (Ours) compared with other 2D style transfer approaches: (a) Content image; (b) AdaIN; (c) AdaAttN; (d) CCPL; (e) MCCNet; (f) HGST (Ours). Please refer to supplementary material for rendered videos.}
  \label{fig:LCSAcompare}
\end{figure*}

Meanwhile, we also compare our proposed Holistic Geometry-preserved Style Transfer model (HGST), with several state-of-the-art 2D stylization methods. As shown in Figure \ref{fig:LCSAcompare}, our method outperforms AdaIN, AdaAttN, and MCCNet by improving temporal consistency while maintaining stylization quality, without exhibiting the blocky pixel artifacts observed in CCPL.

\noindent\textbf{Quantitative Evaluation and Analysis.}  
As shown in \textbf{Table \ref{tab:Quantitative_table}}, We conduct comprehensive quantitative comparisons across three dynamic 4D scenes (\textit{cook\_spinach}, \textit{flame\_salmon\_1}, and \textit{sear\_steak}), evaluating multiple aspects of stylization performance, including imaging quality, aesthetic perception, spatial-temporal consistency, and fidelity to both content and style.

Across all datasets, \textbf{Style4D} consistently achieves top performance in most metrics. For imaging quality, we observe clear improvements in UIQM, Clipiqa+, and Musiq scores, suggesting that our method preserves structural clarity and visual quality more effectively than all baselines. In particular, Style4D achieves a UIQM of 1.9290 on \textit{cook\_spinach} and 1.7529 on \textit{flame\_salmon\_1}, surpassing prior methods such as 4DGS(AdaIN) and 4DStyleGaussian by a large margin.

In terms of aesthetic quality, our method attains the highest Qalign and Musiq-PAQ2PIQ scores across all scenes, indicating better perceptual stylization aligned with artistic intent. For example, on \textit{flame\_salmon\_1}, Style4D yields a Qalign of 3.6030 and a Musiq-PAQ2PIQ of 63.5178, outperforming the second-best baseline by significant margins.

Regarding spatial-temporal consistency, our model shows robust performance with the lowest DISTS and Warp Loss values, confirming its ability to produce temporally stable and coherent results. On \textit{sear\_steak}, Style4D reduces Dists to 0.0108 and Warp Loss to 0.0066, improving both perceptual smoothness and geometric coherence.

Finally, Style4D also excels in content and style consistency. It achieves the best DINO scores, CKDN accuracy, and LPIPS/SSIM metrics across scenes. These results demonstrate our model’s ability to retain scene semantics while applying stylization, balancing content preservation and stylized appearance. For instance, on \textit{cook\_spinach}, it attains a SSIM of 0.7771 and a content LPIPS of 0.1834—clearly outperforming other methods.

\begin{table}[]
\centering
\begin{tabular}{ccc}
\hline
\multirow{2}{*}{Method} & \multicolumn{2}{c}{Frame}           \\ \cline{2-3} 
                        & Stylization Quality & Image Quality \\ \hline
4DGS(AdaIN)             & 11.76\%             & 2.94\%        \\ \hline
4DGS(AdattN)            & 14.70\%             & 8.82\%        \\ \hline
4DStyleGaussian         & 11.76\%             & 17.64\%       \\ \hline
Style4D(Ours)           & \textbf{61.76\%}            &\textbf{ 70.58\%}       \\ \hline
\end{tabular}
\caption{Single-frame image performance: results of user study voting}
\label{us1}
\end{table}

\begin{table}[]
\centering
\resizebox{1.0\linewidth}{!}{%
\begin{tabular}{cccc}
\hline
\multirow{2}{*}{Method} & \multicolumn{3}{c}{Video}                                           \\ \cline{2-4} 
                        & Stylization Quality & Video Quality    & Spatiotemporal Consistency \\ \hline
4DStyleGaussian         & 23.52\%             & 30.12\%          & 44.11\%                    \\
Style4D(Ours)           & \textbf{76.47\%}    & \textbf{69.87\%} & \textbf{55.88\%}           \\ \hline
4DGS(AdaIN)             & 20.58\%             & 11.76\%          & 14.70\%                    \\
Style4D(Ours)           & \textbf{79.41\%}    & \textbf{88.23\%} & \textbf{85.29\%}           \\ \hline
4DGS(AdaAttN)           & 14.70\%             & 17.64\%          & 14.70\%                    \\
Style4D(Ours)           & \textbf{85.29\%}    & \textbf{82.35\%} & \textbf{85.29\%}           \\ \hline
\end{tabular}}
\caption{User study results for video performance evaluation}
\label{us2}
\end{table}

\textbf{Tables~\ref{us1} and~\ref{us2}} present the results of our user study. It can be observed that our method is consistently preferred in terms of both image quality and overall video quality, while simultaneously maintaining strong stylization effects. These findings align well with the quantitative results reported in Table~\ref{tab:Quantitative_table}.

These results collectively confirm that Style4D delivers a strong balance of style fidelity, spatial-temporal coherence, and content preservation, establishing new state-of-the-art performance in 4D stylization.

\noindent\textbf{Further Analyses and Ablation Studies.} We conduct comprehensive ablation studies to evaluate the contribution of each key component in our Style4D framework, validating the necessity and effectiveness of our design. \textbf{Please refer to the supplementary material for detail}.
\section{Conclusion}
\label{sec:conclusion}
We present \textbf{Style4D-Bench}, the first benchmark for 4D stylization, featuring: \textbf{1)} a strong baseline method, \textbf{2)} a unified evaluation protocol covering spatial fidelity, temporal coherence, and multi-view consistency, and \textbf{3)} a curated set of high-resolution dynamic scenes. To establish the baseline, we propose \textbf{Style4D}, a novel 4D stylization framework that combines reliable scene representation, per-Gaussian MLPs for appearance control, and a geometry-preserved stylization module for spatio-temporal consistency. Extensive experiments demonstrate that Style4D outperforms existing methods, delivering high-quality stylization with stable dynamics and coherent multi-view rendering. We expect Style4D-Bench to promote future research in 4D stylized scene synthesis. In future work, we plan to further improve the quality of stylization and broader user-controllable style manipulation for interactive applications.

\textbf{Limitation.} Although Style4D achieves high-quality stylization with spatial-temporal consistency, it still faces limitations. The training process can be time-consuming due to the multi-stage design and per-Gaussian MLPs. Additionally, the current framework focuses on a fixed style per scene; supporting rapid style switching or region-specific stylization remains an open challenge.

\section*{Acknowledgement}  The computational resources are supported by SongShan
Lake HPC Center (SSL-HPC) in Great Bay University. This
work was also supported by Guangdong Research Team
for Communication and Sensing Integrated with Intelligent
Computing (Project No. 2024KCXTD047).

{
    \small
    \bibliographystyle{ieeenat_fullname}
    \bibliography{main}
}
\clearpage
\appendix

\section{Implementation and Network Details}
\paragraph{\textbf{4DGS Representation.}} 
We follow the training configurations described in the 4DGS \cite{wu20244d} paper for this component. Stylization experiments are conducted on the real-world \texttt{Neu3D} \cite{li2022neural} dataset, which contains six dynamic scenes. Each input image has a resolution of $1352 \times 1014$, and the reconstruction is performed at the same resolution, with each video comprising 300 frames. The initial point cloud for each scene is generated using the official initialization code provided by 4DGS \footnote{https://github.com/hustvl/4DGaussians}. After initialization, each scene is downsampled to approximately 3--4k Gaussian points.

\paragraph{\textbf{Style Gaussian Representation.}} 
Each tiny MLP receives the time step $t$ and intersection depth as inputs. The network consists of four hidden units and outputs RGB deltas along with opacity values (4-dimensional output). The learning rate is scheduled with $\text{lr}_{\text{init}} = 0.0001$, $\text{lr}_{\text{final}} = 0.00001$, and a delay multiplier $\text{lr}_{\text{delay\_mult}} = 0.02$.

\paragraph{\textbf{Holistic Geometry-preserved Style Transfer Module.}} 
For this module, we use images from the COCO2014 dataset \cite{lin2014microsoft} as content inputs and style images sampled from WikiArt. The model is evaluated on Neu3D scenes to demonstrate strong generalization ability. During training, images are cropped to $256 \times 256$ resolution. The overall loss function is defined as:

\begin{equation}
\begin{aligned}
\mathcal{L}_{\text{total}} = 
\lambda_{\text{consistency}} \mathcal{L}_{\text{consistency}} +
\lambda_{\text{style}} \mathcal{L}_{\text{style}} +
\lambda_{\text{id}} \mathcal{L}_{\text{id}} 
\\
+ \lambda_{\text{illum}} \mathcal{L}_{\text{illum}} +
\lambda_{\text{ins}} \mathcal{L}_{\text{ins}},
\end{aligned}
\end{equation}

\noindent
where the weights are set as follows: 
$\lambda_{\text{consistency}} = 3$, 
$\lambda_{\text{style}} = 18$, 
$\lambda_{\text{id}} = 7$, 
$\lambda_{\text{illum}} = 10^{-5}$, 
$\lambda_{\text{ins}} = 1$.

1) The style perceptual loss $\mathcal{L}_{style}$ minimizes the style differences between the generated image \( I_{cs} \) and the style image \( I_s \) by comparing the mean and variance of features extracted from each layer of a pre-trained VGG19. Formally, the loss is expressed as:

\begin{equation}
\begin{aligned}
\mathcal{L}_{style} = \sum_{l} & \left( \| \mu_l(I_{cs}) - \mu_l(I_s) \|_2^2 \right. \\
& \left. + \| \sigma_l(I_{cs}) - \sigma_l(I_s) \|_2^2 \right),
\end{aligned}
\end{equation}

where $\mu_l(\cdot)$ and $\sigma_l(\cdot)$ denote the mean and standard deviation computed over the spatial dimensions of the feature map at layer $l$. Given the feature map $F_l(I) \in \mathbb{R}^{C_l \times H_l \times W_l}$, these statistics are defined as:

\begin{equation}
\begin{aligned}
\mu_l(I) = \frac{1}{H_l W_l} \sum_{h=1}^{H_l} \sum_{w=1}^{W_l} F_l(I)_{:, h, w}, \\
\quad
\sigma_l(I) = \sqrt{ \frac{1}{H_l W_l} \sum_{h=1}^{H_l} \sum_{w=1}^{W_l} \left( F_l(I)_{:, h, w} - \mu_l(I) \right)^2 },
\end{aligned}
\end{equation}

where the operations are computed channel-wise.

\begin{figure*}[h]
  \centering
  \includegraphics[width=1\textwidth]{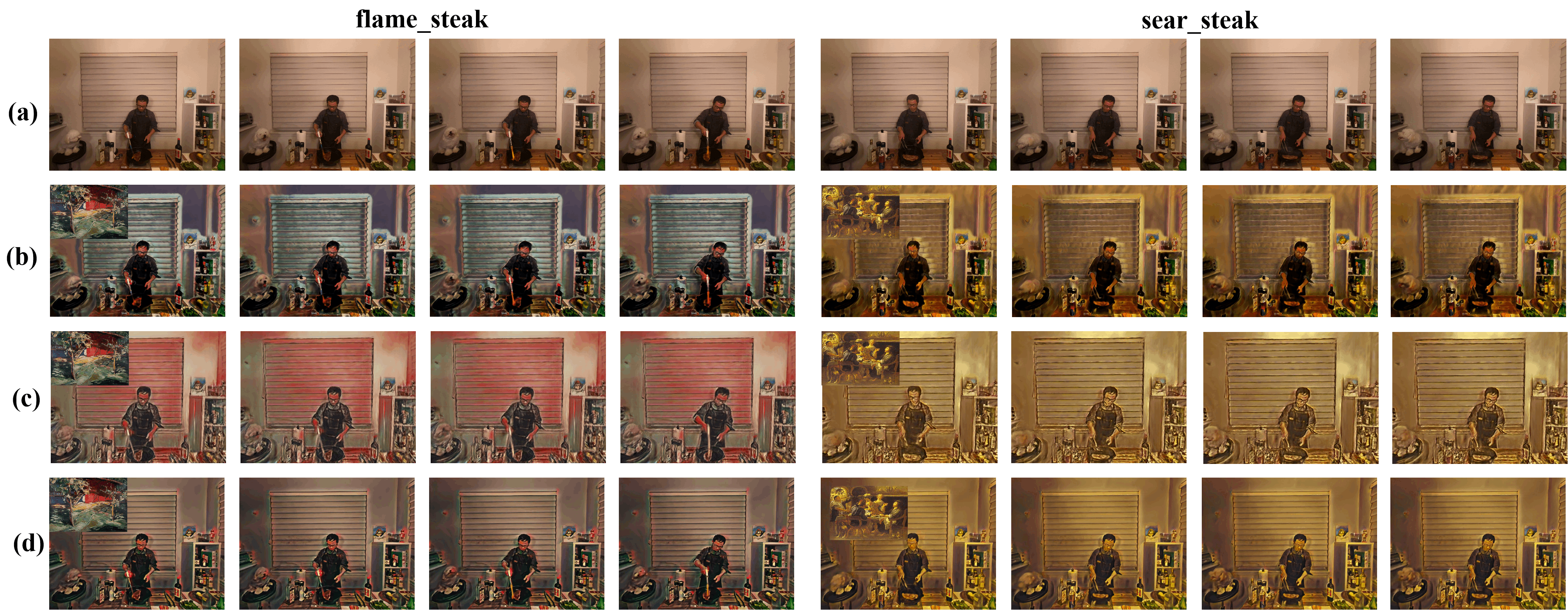}
  \caption{Comparison of stylization results by different methods: (a) Original images (before style transfer), (b) 4DGS with AdaIN, (c) 4DGS with AdaAttN, (d) Style4D(Ours).}
  \label{fig:Qualitative_results}
\end{figure*}

\begin{figure*}[h]
  \centering
  \includegraphics[width=0.7\textwidth]{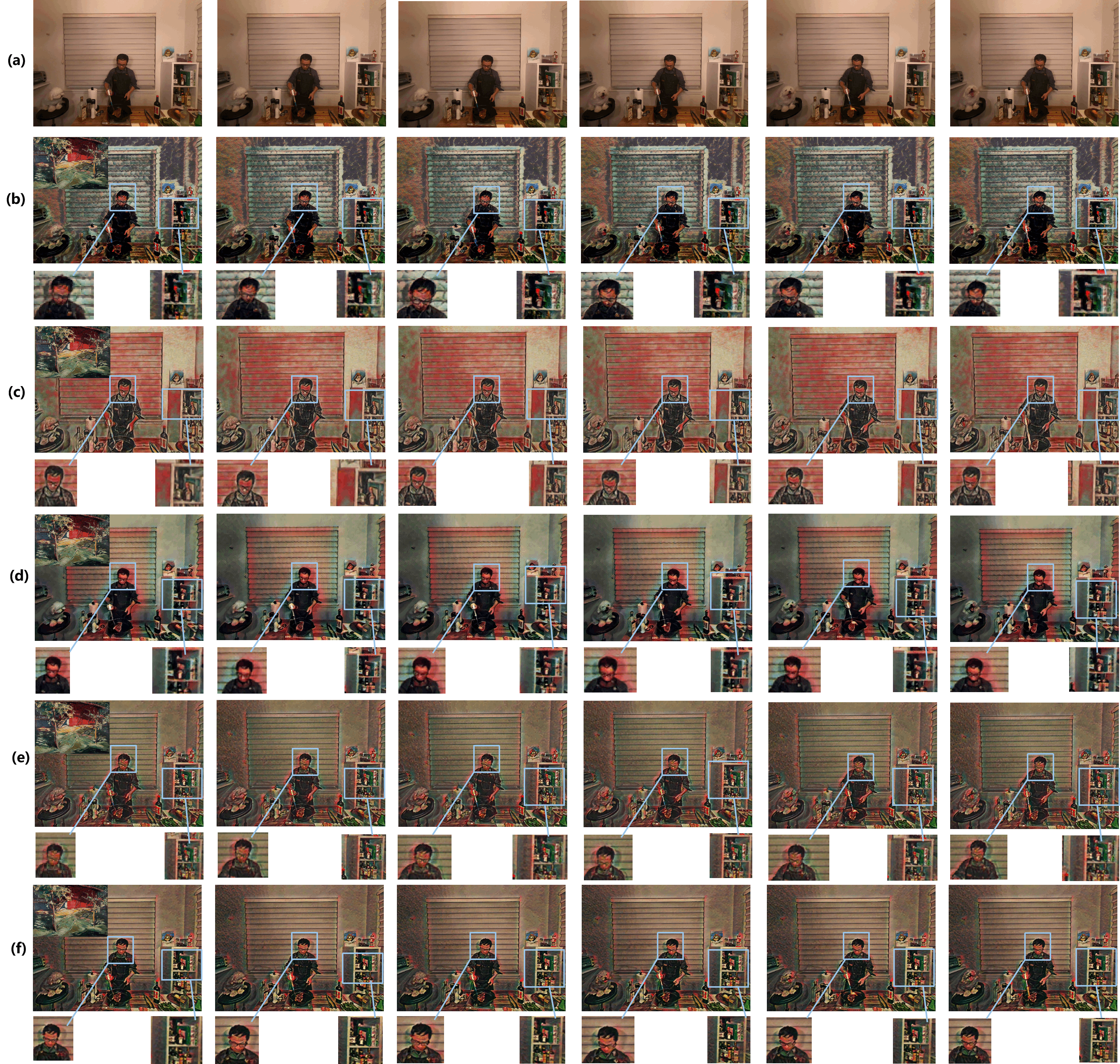}
  \caption{Visualization of HGST stylization method (Ours) compared with other 2D style transfer approaches: (a) Content image; (b) AdaIN; (c) AdaAttN; (d) CCPL; (e) MCCNet; (f) HGST (Ours).}
  \label{fig:HGST_rev}
\end{figure*}

2) The identity loss $\mathcal{L}_{id}$ helps to preserve the content structure while maintaining the richness of the style patterns:
\begin{equation}
\mathcal{L}_{id} = \| I_{cc} - I_c \|^2 + \| I_{ss} - I_s \|^2
\end{equation}
where \( I_{cc} \) and \( I_{ss} \) are the generated results using natural images and paintings as content and style images, respectively.

3) Illumination Loss \( \mathcal{L}_{illum} \):
Illumination loss addresses the flickering effect caused by illumination variations in video sequences. It is defined as:
\begin{equation}
\mathcal{L}_{illum} = \| G(I_c, I_s) - G(I_c + \epsilon, I_s) \|^2
\end{equation}
where \( G(\cdot) \) is the generation function, and \( \epsilon \sim N(0, \sigma^2 I) \) represents random Gaussian noise.

4) Inner Channel Similarity Loss \( \mathcal{L}_{ins} \):
This loss strengthens the consistency of generated features within each channel, ensuring that there are no disharmonious areas:
\begin{equation}
\mathcal{L}_{ins} = \sum_{c=1}^{C} \mathrm{Inner}_{c,i}
\end{equation}
where \( \mathrm{Inner}_{c,i} \) is the inner similarity defined as:
\begin{equation}
\mathrm{Inner}_{c,i} = \arg\min_{i} \sum_{j=1}^{h \times w} \left( 1 - \frac{f_i \cdot f_j}{\| f \|_2^2} \right)
\end{equation}

where \( f \) represents the generated features, and \( h \times w \) is their resolution.

\section{Additional comparative experiments}
\subsection{Quantitative Results}
\textbf{Metrics.} The goal of stylization is to transform object edges and texture details to match the style of a reference image while preserving the overall structural integrity of the original image. To evaluate the structural similarity with the original image, we employ SSIM and LPIPS metrics. To quantitatively assess the degree of stylization, we use a style loss computed as follows: features are extracted from five layers of a pretrained VGG network \cite{simonyan2014very} for both the stylized image $I_{cs}$ and the reference style image $I_s$. The style loss is defined as the sum of mean squared errors (MSE) between the Gram matrices of corresponding feature layers of $I_{cs}$ and $I_s$. This accumulated style loss serves as a measure of the stylistic discrepancy between the generated image and the target style. The Gram matrix is computed as follows: Given the feature map at a certain layer $F \in \mathbb{R}^{C \times H \times W}$, where $C$ denotes the number of channels, and $H$ and $W$ denote the height and width of the feature map, respectively. We first reshape $F$ into a matrix $F' \in \mathbb{R}^{C \times (H \times W)}$, then compute the Gram matrix $G \in \mathbb{R}^{C \times C}$ as
\begin{equation}
G_{ij} = \frac{1}{C \times H \times W} \sum_{k=1}^{H} \sum_{l=1}^{W} F_{i,k,l} \times F_{j,k,l},
\end{equation}
where $G_{ij}$ represents the correlation between the $i$-th and $j$-th channel features, capturing the style information encoded in that layer.

\begin{table}[]
\centering
\resizebox{0.5\textwidth}{!}{
\begin{tabular}{cccccc}
\hline
Method    & AdaIN    & AdaAttN  & CCPL     & MCCNet   & HGST(Ours) \\ \hline
Warp Loss↓ & 0.045824 & 0.031671 & 0.013301 & 0.021858 & 0.021068   \\ \hline
\end{tabular}}
\caption{Comparison of Warp Loss for Video Stylization Results Across Different Methods.}
\label{tab:warploss}
\end{table}

\begin{table}[]
\centering
\begin{tabular}{cccc}
\hline
Method          & Dataset                           & SSIM↑           & LPIPS↓          \\ \hline
4DGS(AdaIN)     & \multirow{4}{*}{sear\_steak}      & 0.6000          & 0.2987          \\
4DGS(AdaAttN)   &                                   & 0.5153          & 0.4768          \\
4DStyleGaussian &                                   & 0.6557          & 0.4819          \\
Style4D(Ours)   &                                   & \textbf{0.7503} & \textbf{0.2146} \\ \hline
4DGS(AdaIN)     & \multirow{4}{*}{cook\_spinach}    & 0.4763          & 0.4898          \\
4DGS(AdaAttN)   &                                   & 0.6444          & 0.2841          \\
4DStyleGaussian &                                   & 0.7646          & 0.2159          \\
Style4D(Ours)   &                                   & \textbf{0.7771} & \textbf{0.1834} \\ \hline
4DGS(AdaIN)     & \multirow{4}{*}{flame\_salmon\_1} & 0.5201          & 0.4030          \\
4DGS(AdaAttN)   &                                   & 0.6054          & 0.2944          \\
4DStyleGaussian &                                   & 0.5081          & 0.6051          \\
Style4D(Ours)   &                                   & \textbf{0.6963} & \textbf{0.2704} \\ \hline
\end{tabular}
\caption{Quantitative comparisons of Style4D against other methods.}
\label{tab:Quantitative_table}
\end{table}

Table~\ref{tab:Quantitative_table} presents a quantitative comparison between our method and other stylization approaches on rendered stylized videos from novel test viewpoints. The results demonstrate that our method effectively preserves the underlying scene structure while achieving strong stylization performance.

Since both our method and 4DStyleGaussian exhibit strong visual consistency, we employ style loss as a quantitative metric to better evaluate stylization quality. Table~\ref{tab:Quantitative_table2} presents the style loss between the stylized results of our method, 4DStyleGaussian, and the reference style image. The results indicate that our method achieves closer adherence to the reference style while effectively preserving the original scene structure.

\begin{table}[]
\centering
\begin{tabular}{ccc}
\hline
Method          & Dataset                           & Style Loss↓       \\ \hline
4DStyleGaussian & \multirow{2}{*}{sear\_steak}      & 0.006816          \\
Style4D(Ours)   &                                   & \textbf{0.005687} \\ \hline
4DStyleGaussian & \multirow{2}{*}{cook\_spinach}    & 0.008329          \\
Style4D(Ours)   &                                   & \textbf{0.006123} \\ \hline
4DStyleGaussian & \multirow{2}{*}{flame\_salmon\_1} & 0.007387          \\
Style4D(Ours)   &                                   & \textbf{0.006364} \\ \hline
\end{tabular}
\caption{Quantitative Comparison of Stylization Performance.}
\label{tab:Quantitative_table2}
\end{table}

\begin{figure*}[h]
  \centering
  \includegraphics[width=0.7\textwidth]{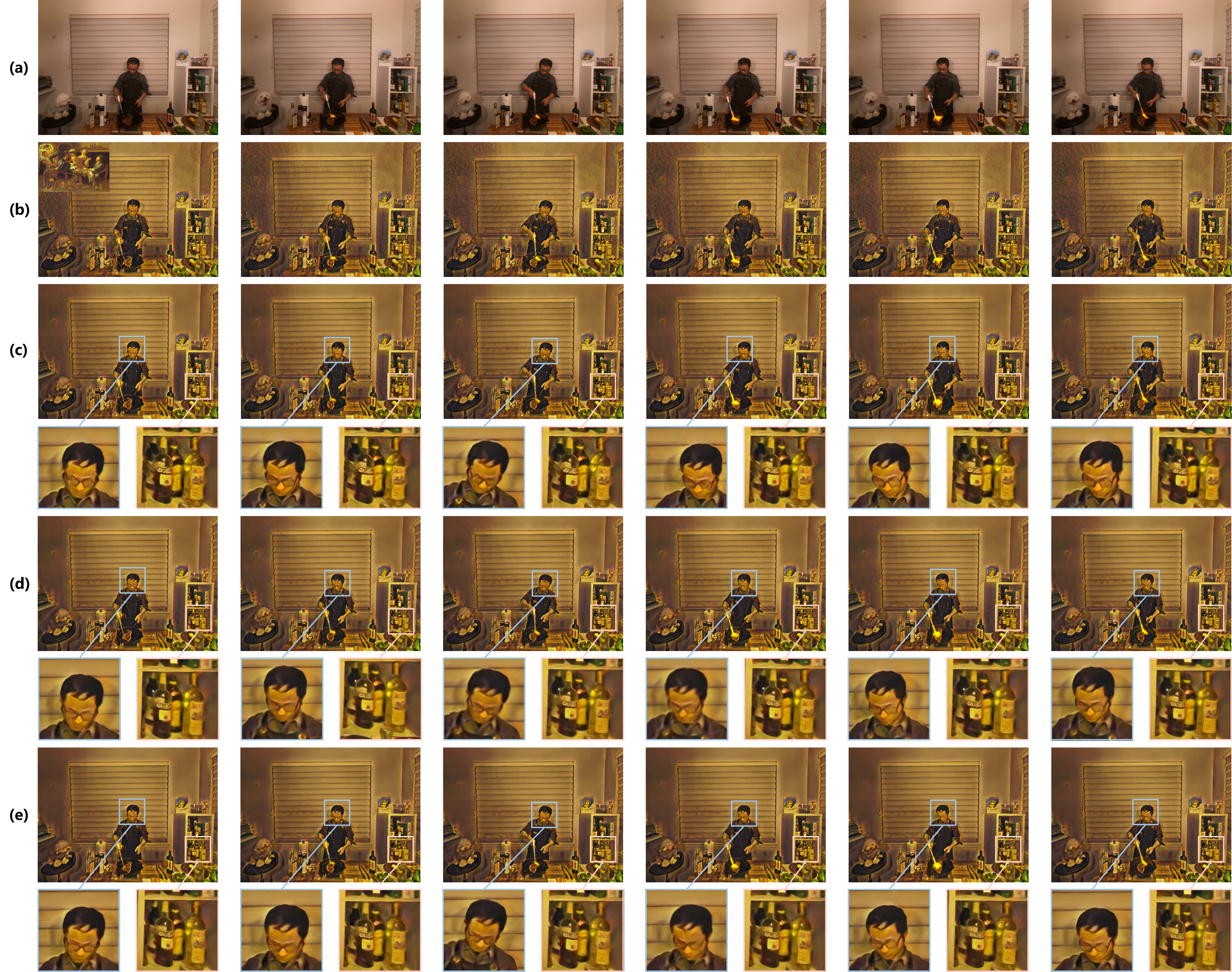}
  \caption{Ablation study: (a) Original scene images, (b) Stylized images, (c) 4DGS trained with  stylized images, (d) 4DGS trained under our two-stage scheme without lightweight MLPs for gaussians, (e) Style4D (Ours). Our method better preserves fine-grained details such as facial structures and accessories while minimizing the artifacts.}
  \label{fig:Ablation_static}
\end{figure*}
\begin{figure*}[h]
  \centering
  \includegraphics[width=1\textwidth]{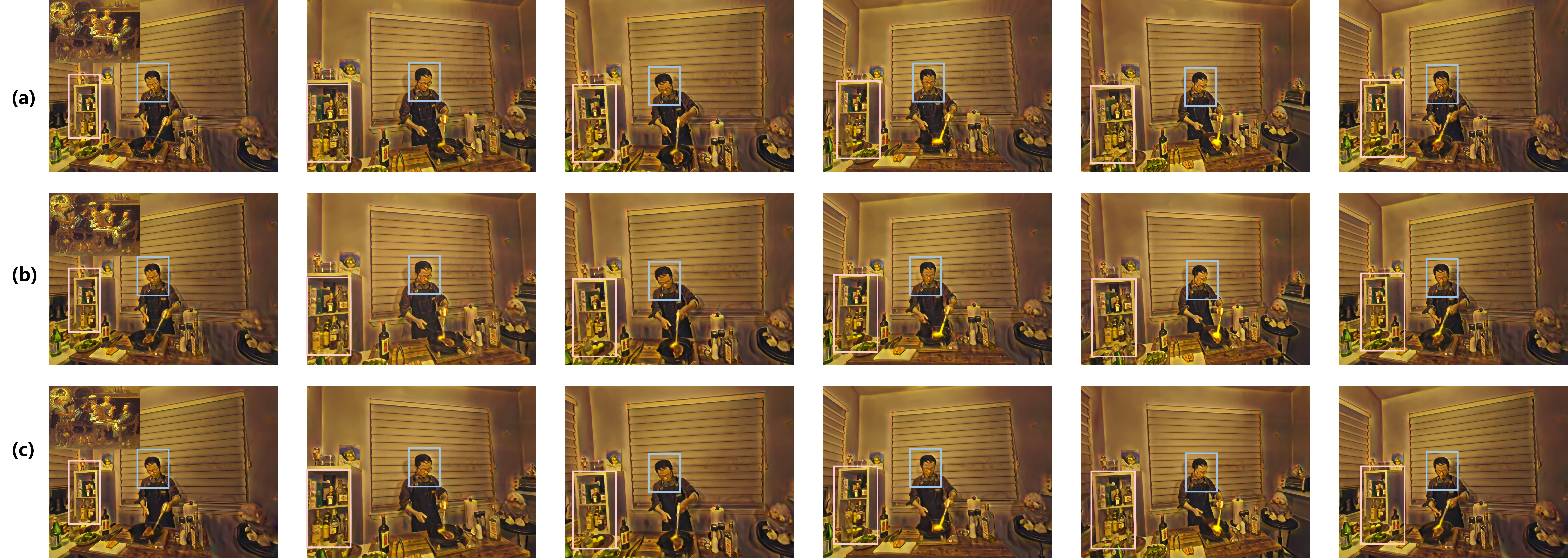}
  \caption{Ablation study with moving views: (a) 4DGS trained with  stylized images, (b) 4DGS trained under our two-stage scheme without lightweight MLPs for gaussians, (c) Style4D (Ours). Our method achieves a better multi-view consistencys.}
  \label{fig:Ablation_dynamic}
\end{figure*}

\textbf{Holistic Geometry-preserved Style Transfer(HGST)} To quantitatively evaluate the temporal consistency of different stylization methods, we employ the warp loss metric. We estimate the optical flow between consecutive frames using the RAFT method \cite{teed2020raft}. Let $I_t$ and $I_{t+1}$ denote image frames at time steps $t$ and $t+1$, respectively, and $\mathbf{F}_{t \to t+1}$ be the optical flow field mapping pixel coordinates from frame $t$ to frame $t+1$. Using this flow, the subsequent frame $I_{t+1}$ is spatially warped to produce a transformed image $\hat{I}_{t+1}$ aligned with $I_t$:

\begin{equation}
\hat{I}_{t+1}(\mathbf{x}) = I_{t+1} \big( \mathbf{x} + \mathbf{F}_{t \to t+1}(\mathbf{x}) \big),
\end{equation}

where $\mathbf{x}$ denotes pixel coordinates. The warp loss between two frames is then defined as the pixel-wise difference between $\hat{I}_{t+1}$ and $I_t$:

\begin{equation}
\mathcal{L}_{warp}^t = \frac{1}{HW} \sum_{\mathbf{x}} \left| \hat{I}_{t+1}(\mathbf{x}) - I_t(\mathbf{x}) \right|,
\end{equation}

where $H$ and $W$ are the height and width of the image, respectively. The overall warp loss for a video sequence is computed as the average over all consecutive frame pairs:

\begin{equation}
\mathcal{L}_{warp} = \frac{1}{T-1} \sum_{t=1}^{T-1} \mathcal{L}_{warp}^t,
\end{equation}

where $T$ is the total number of frames in the video. This metric reflects the temporal coherence across video frames.

Table~\ref{tab:warploss} presents a comparison of temporal consistency among our proposed HGST method and other stylization approaches in the video stylization task. Our method demonstrates superior consistency compared to most baselines, slightly trailing behind CCPL. However, CCPL achieves this consistency improvement at the cost of sacrificing local detail fidelity.

\subsection{Qualitative results}

Figure~\ref{fig:Qualitative_results} presents qualitative comparisons of rendered images from novel test viewpoints between our method and several baselines. Although AdaIN \cite{huang2017arbitrary} and AdaAttN \cite{liu2021adaattn} achieve stronger stylization effects, they substantially distort the original scene structure and introduce significant artifacts and noise. In contrast, our method (d) achieves a high degree of style transfer while faithfully preserving the scene structure and fine facial details.

\textbf{Holistic Geometry-preserved Style Transfer(HGST)} Due to space constraints in the main text, we provide here a detailed comparison of our 2D stylization method HGST against other state-of-the-art 2D stylization approaches, as well as a corrected visualization of our method’s results. As shown in Figure~\ref{fig:HGST_rev}, while AdaIN (b) and AdaAttN (c) exhibit strong stylization effects, they suffer from poor temporal consistency in video sequences. MCCNet (e) inadequately preserves local detail features, resulting in noticeable background flickering and blurred shadows between subjects and backgrounds. Although CCPL maintains strong temporal consistency, it compromises local detail, leading to heavily blurred facial regions and numerous hole-like artifacts in the background. In contrast, our method effectively balances stylization quality with preservation of original image structure and temporal coherence.

\begin{table}[]
\centering
\resizebox{0.5\textwidth}{!}{
\begin{tabular}{ccccccc}
\hline
\multirow{2}{*}{Method} & \multicolumn{6}{c}{Dataset}                                                                                \\ \cline{2-7} 
                        & flame\_steak   & sear\_steak    & flame\_salmon\_1 & coffee\_martini & cook\_spinach  & cut\_roasted\_beef \\ \hline
4DGS                    & 32.02          & 31.75          & 28.69            & 28.68           & 32.06          & 32.55              \\
ours                    & \textbf{32.85} & \textbf{32.60} & \textbf{28.89}   & \textbf{28.87}  & \textbf{32.28} & \textbf{32.81}     \\ \hline
\end{tabular}
}
\caption{Comparison of PSNR between Our Framework and 4DGS on Various Datasets.}
\label{tab:psnr}
\end{table}

\begin{figure}[h]
  \centering
  \includegraphics[width=0.45\textwidth]{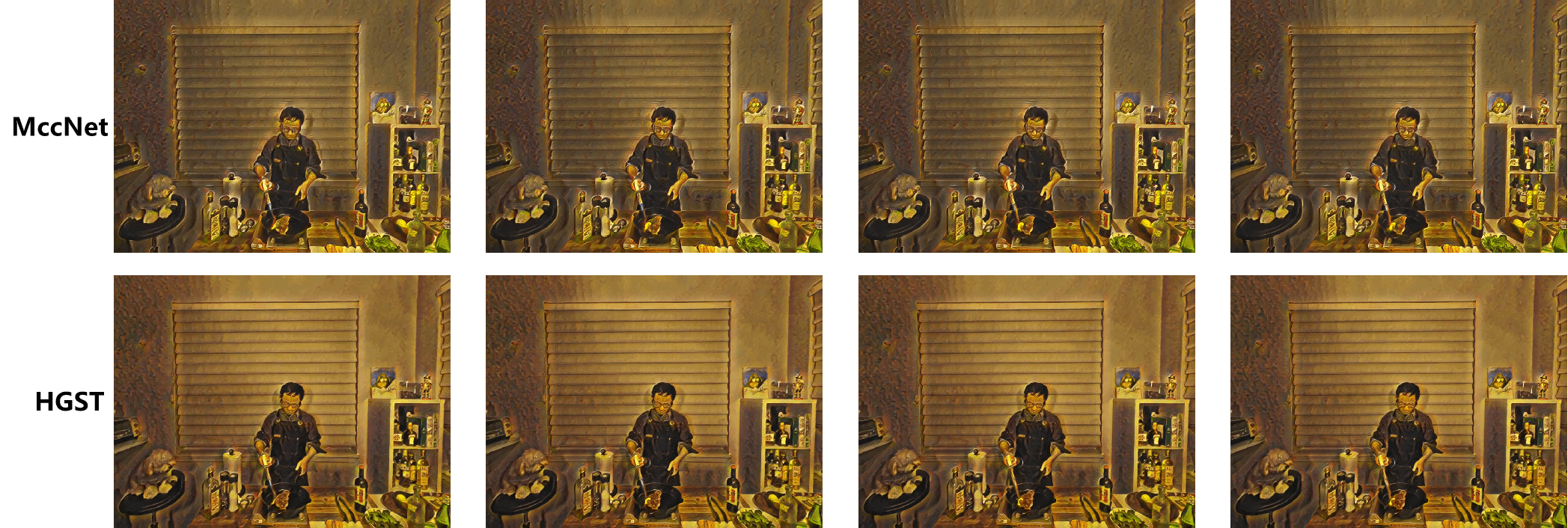}
  \caption{Ablation study of HGST.}
  \label{fig:LCSA}
\end{figure}

\section{Ablation Studies}
\label{sec:experiments}

We conduct comprehensive ablation studies to evaluate the contribution of each key component in our Style4D framework, validating the necessity and effectiveness of our design.

\textbf{Compared with Direct 4DGS Training on Stylized Images.}  
We first investigate the baseline of directly training 4D Gaussian Splatting (4DGS) on stylized images, without decoupling geometry reconstruction from stylization and lightweight MLPs for stylization. As illustrated in the row c in Figure \ref{fig:Ablation_static} , this baseline produces substantially degraded results characterized by temporal flickering and spatial artifacts. Such outcomes confirm the necessity of our proposed two-stage training pipeline, which disentangles geometry learning (coarse and fine stages) from stylization, and enhances the representation ability for each gaussian, thereby ensuring a stable 4D scene structure and enabling flexible and high-quality style transfer.

\textbf{Effectiveness of Per-Gaussian MLPs.}  To assess the impact of the per-Gaussian MLPs introduced in the style stage, we compare our full model against a variant that directly optimizes Gaussian parameters without the use of MLPs. As shown in the row d in Figure \ref{fig:Ablation_static}, removing the MLPs results in noticeable degradation of overall stylization quality and a loss of fine-grained texture details. The per-Gaussian MLPs facilitate spatial-temporal modulation of appearance, providing precise control over style evolution across both time and viewpoint, which is critical for achieving smooth temporal transitions and multi-view consistency.

\begin{figure*}[!p]
  \centering
  \includegraphics[width=1\textwidth]{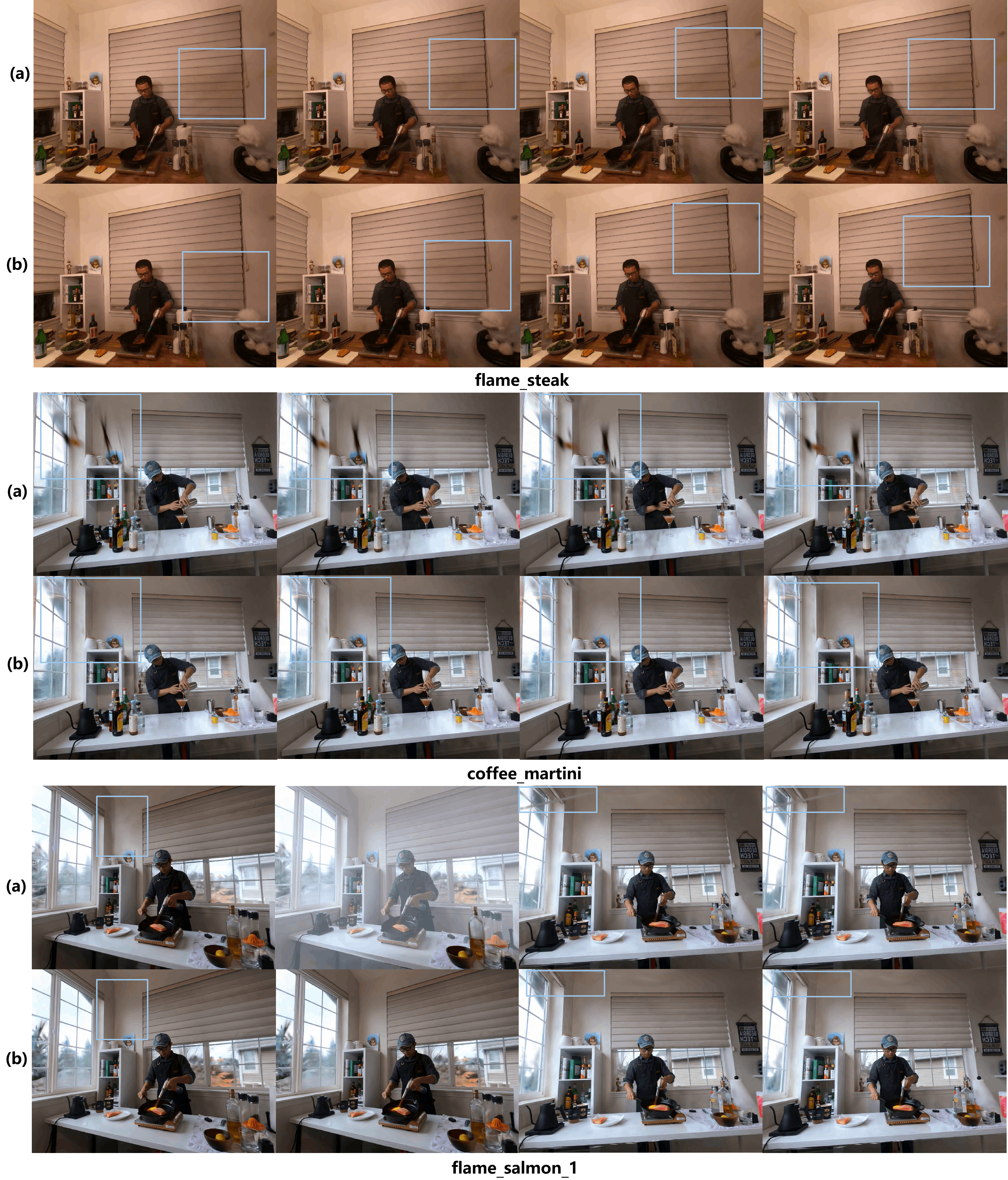}
  \caption{Reconstruction Results under Moving Viewpoints on flame\_steak, coffee\_martini, and flame\_salmon\_1 Datasets: (a) 4DGS (b) Ours.}
  \label{fig:GSstyle_dynerf_1}
\end{figure*}

\begin{figure*}[!p]
  \centering
  \includegraphics[width=1\textwidth]{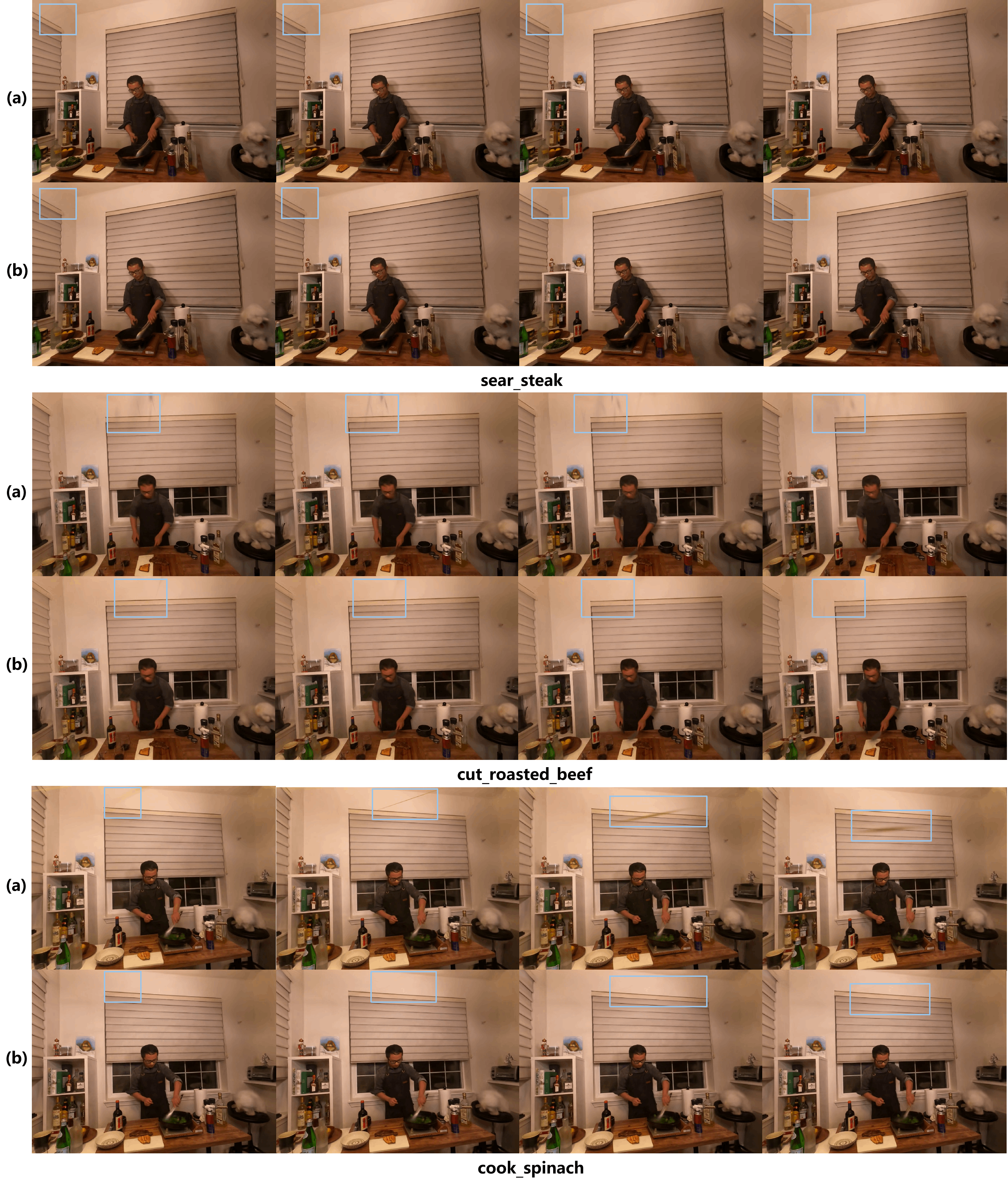}
  \caption{Reconstruction Results under Moving Viewpoints on sear\_steak, cut\_roasted\_beef, and cook\_spinach Datasets: (a) 4DGS (b) Ours.}
  \label{fig:GSstyle_dynerf_2}
\end{figure*}

To further validate the effectiveness of our proposed style Gaussian representation, we conduct training and evaluation on the original datasets and compare with 4DGS. As shown in Table \ref{tab:psnr} and Figure \ref{fig:GSstyle_dynerf_1}, \ref{fig:GSstyle_dynerf_2}, our Gaussian representation outperforms 4DGS, significantly reducing artifacts and blurriness in the synthesized novel views.

\textbf{Effectiveness of Geometry-guided Initialization.}  
We further evaluate if the learned geometry from earlier stages could help achieving a better view-consistent stylization result by comparing training 4DGS with stylized images with and without geometry-informed priors. Figure \ref{fig:Ablation_static} demonstrates that training stylized 4DGS from scratch will lead to inferior visual fidelity and structural artifacts, confirming that geometry-guided initialization provides essential structural priors.

\textbf{Effectiveness of Holistic Geometry-preserved Style Transfer.}
As illustrated in Figure \ref{fig:LCSA}, our proposed HGST model significantly improves spatial-temporal consistency compared to the original MCCNet while maintaining stylization quality. Flickering artifacts in the video are substantially reduced.

\begin{figure*}[!p]
  \centering
  \includegraphics[width=1\textwidth]{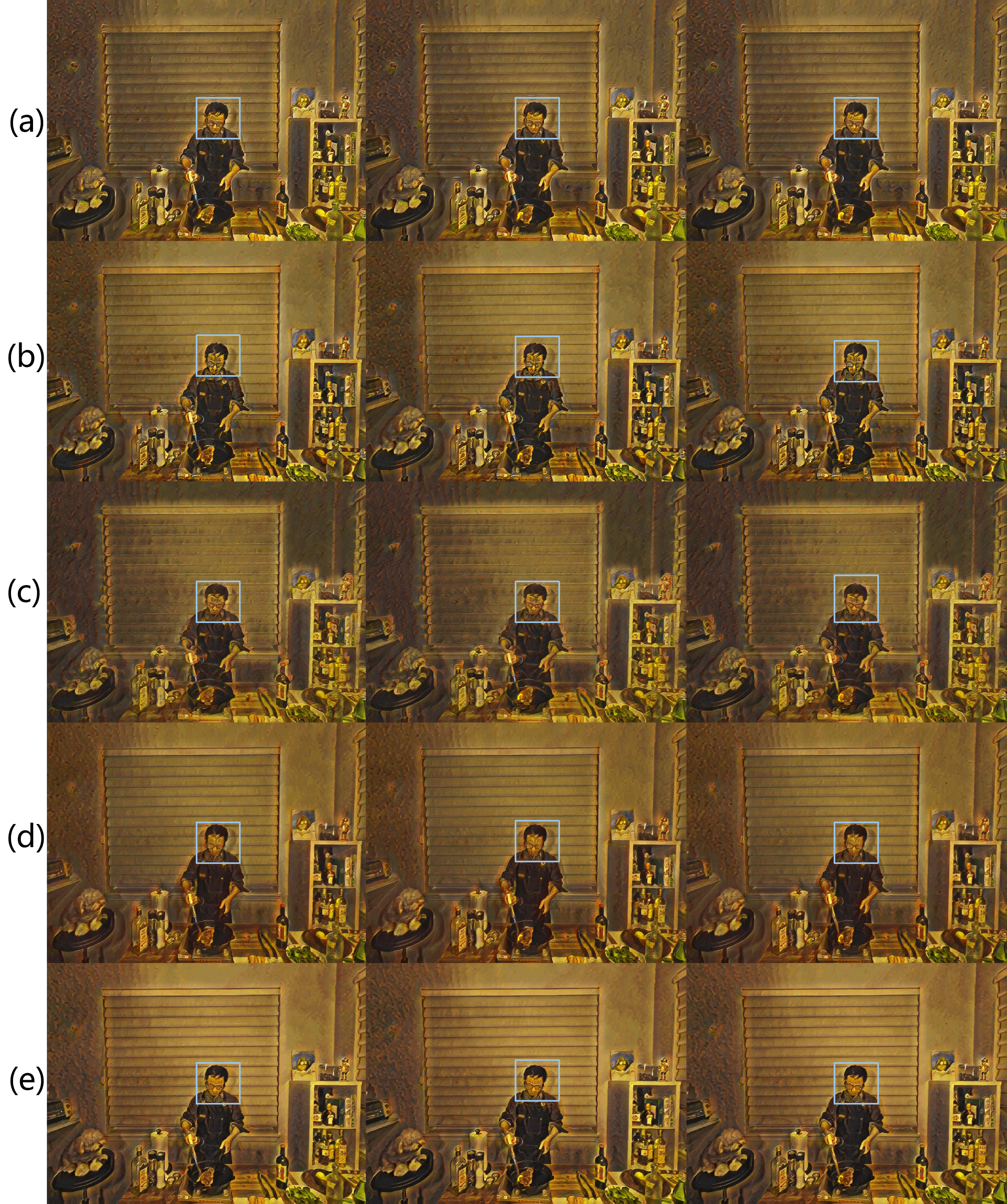}
  \caption{Ablation Study of the Holistic Geometry-preserved Style Transfer Module: (a) MCCNet (b) MCCNet with CCPL Loss (c) MCCNet with Content Loss (d) MCCNet with LCL Loss (e) HGST (Ours).}
  \label{fig:full_ablation_HGST}
\end{figure*}

The full ablation study of the proposed HGST module is illustrated in Figure~\ref{fig:full_ablation_HGST}. Although (a) MCCNet, our baseline, preserves spatiotemporal consistency during stylization, it still suffers from severe flickering and exhibits poor detail consistency, especially on high-resolution frames. To evaluate the effectiveness of our proposed local consistency loss (LCL) and content loss, we conducted the following experiments: (b) incorporates the CCPL contrastive loss into MCCNet. While it introduces some structural consistency, it also leads to noticeable hollow artifacts around the face and adjacent regions. (c) integrates the content loss into MCCNet, which improves overall consistency, but results in blurring of physical shapes and boundaries. (d) adds our LCL loss on top of MCCNet, which enhances local consistency and alleviates hollow artifacts to some extent; however, some artifacts remain, and flickering is still apparent in the background curtain region. Finally, (e) presents our complete Holistic Geometry-preserved Style Transfer (HGST) module. Compared to all prior variants, our method achieves significantly better global and local consistency, while substantially reducing temporal flickering artifacts.

\end{document}